\documentclass[10pt,twocolumn,letterpaper]{article}

\usepackage{cvpr}
\usepackage{times}
\usepackage{epsfig}
\usepackage{graphicx}
\usepackage{amsmath}
\usepackage{amssymb}
\usepackage{hyperref}
\usepackage{subcaption}
\usepackage{multirow}
\usepackage{multicol}

\cvprfinalcopy 

\ifcvprfinal\fi

\newcommand{\minisection}[1]{\vspace{0.04in} \noindent {\bf #1}\ \ }

\begin{document}

\title{Mix and match networks: encoder-decoder alignment for zero-pair image translation}
\author{Yaxing Wang, Joost van de Weijer, Luis Herranz \\
Computer Vision Center, Universitat Aut\`onoma de Barcelona\\
Barcelona, Spain\\
{\tt\small \{wang,joost,lherranz\}@cvc.uab.es}
}

\maketitle

\begin{abstract}
We address the problem of image translation between domains or modalities for which no direct paired data is available (i.e. zero-pair translation). We propose mix and match networks, based on multiple encoders and decoders aligned in such a way that other encoder-decoder pairs can be composed at test time to perform unseen image translation tasks between domains or modalities for which explicit paired samples were not seen during training. We study the impact of autoencoders, side information and losses in improving the alignment and transferability of trained pairwise translation models to unseen translations.
We show our approach is scalable and can perform colorization and style transfer between unseen combinations of domains.
We evaluate our system in a challenging cross-modal setting where semantic segmentation is estimated from depth images, without explicit access to any depth-semantic segmentation training pairs. 
Our model outperforms baselines based on pix2pix and CycleGAN models.
\end{abstract}

\section{Introduction}
Image-to-image translations (or simply image translations) are an integral part of many computer vision systems. They include transformations between different modalities, such as from RGB to depth~\cite{liu2016learning}, or domains, such as luminance to color images~\cite{zhang2016colorful}, horses to zebras~\cite{zhu2017unpaired}, or editing operations such as artistic style changes~\cite{gatys2016image}. These mappings can also  include 2D label representations such as semantic segmentations~\cite{long2015fully} or surface normals~\cite{eigen2015predicting}. Deep networks have shown excellent results in learning models to perform image translations between different domains and modalities~\cite{badrinarayanan2015segnet,isola2016image,long2015fully}. These systems are typically trained with pairs of matching images between domains, e.g. an RGB image and its corresponding depth image. 

\begin{figure}[t]
\centering
\includegraphics[width=0.99\columnwidth]{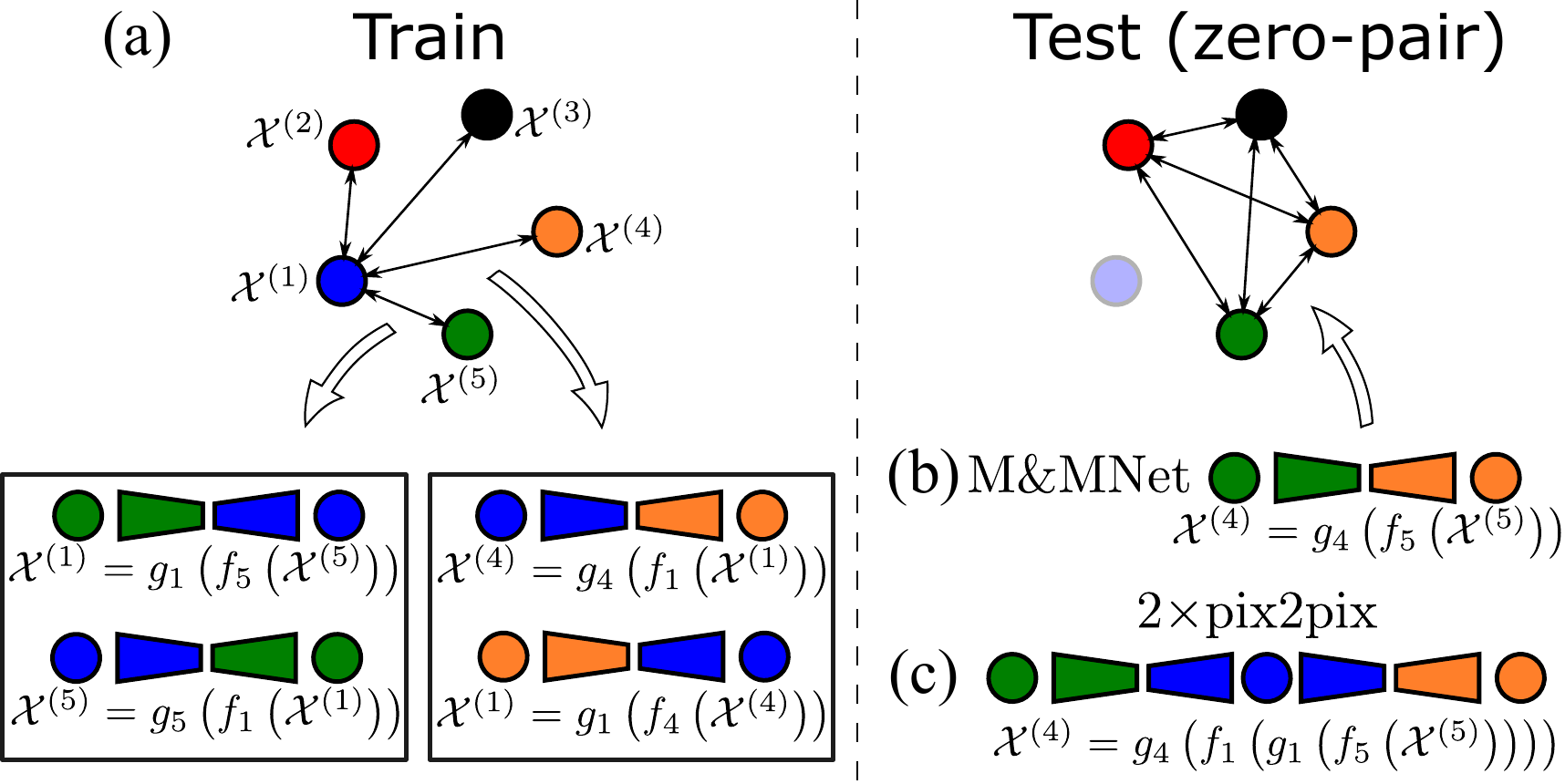}
\caption{\textit{Zero-pair image translation}: (a) given a set of domains or modalities (circles) for which paired training data is available, the objective is to evaluate zero-pair translations. Translations are implemented as aligned encoder-decoder networks. (b) \textit{Mix and match networks} do not require retraining on unseen transformations, in contrast to unpaired translation alternatives (e.g. CycleGAN \cite{zhu2017unpaired}). (c) Two cascaded paired translations (e.g. 2$\times$pix2pix~\cite{isola2016image}) require explicit translation to an intermediate domain. Better seen in color.}
\label{fig_zero-pair_translation}
\end{figure}

Image translation methods, which transfer images from one domain to another, are often based on encoder-decoder frameworks~\cite{badrinarayanan2015segnet,isola2016image,long2015fully,zhu2017unpaired}. In these approaches, an encoder network maps the input image from domain A to a continuous vector in a
latent space. From this latent representation the decoder generates an image in domain B. The latent representation is typically much smaller than the original image size, thereby forcing the network to learn to efficiently compress the information from domain A which is relevant for domain B into the latent representation. Encoder-decoder networks are trained end-to-end by providing the network with matching pairs from both domains or modalities. An example could be learning a mapping from RGB to depth~\cite{liu2016learning}. Other applications include semantic segmentation ~\cite{badrinarayanan2015segnet} and image restoration~\cite{mao2016image}.

In this paper we introduce \textit{zero-pair image translation}: a new setting for testing image translations which involves evaluating on \textit{unseen} translations, i.e. no matching image or dataset pairs are available during training (see Figure~\ref{fig_zero-pair_translation}a). Note that this setting is different from unpaired image translation~\cite{zhu2017unpaired,kim2017learning,liu2017unsupervised}, which is evaluated on the same paired domains seen during training.

We also propose \textit{mix and match networks}, an approach that addresses zero-pair image translation by seeking  alignment between encoders and decoders via their latent spaces. An unseen translation between two domains is performed by simply concatenating the input domain encoder and the output domain decoder (see Figure~\ref{fig_zero-pair_translation}b). We study several techniques that can improve this alignment, including the usage of autoencoders, latent space consistency losses and the usage of pooling indices as side information to guide the reconstruction of spatial structure. We evaluate this approach in a challenging cross-modal task, where we perform zero-pair depth to semantic segmentation translation, using only RGB to depth and RGB to semantic segmentation pairs during training. 

Finally, we show that aligned encoder-decoder networks also have advantages in domains with unpaired data. In this case, we show that mix and match networks scale better with the number of domains, since they are not required to learn all pairwise image translation networks (i.e. scales linearly instead of quadratically). The code is available at http://github.com/yaxingwang/Mix-and-match-networks

\section{Related Work}
\paragraph{Image translation}
Recently, generic encoder-decoder architectures have achieved impressive results in a wide range of transformations between images. Isola et al.~\cite{isola2016image} trained from pairs of input and output images to learn a variety of image translations (e.g. color, style), using an adversarial loss. 
These models require paired training data to be available (i.e. \textit{paired} image translation). Various works extended this idea to the case where no explicit input-output image pairs are available (\textit{unpaired} image translation), using the idea of cyclic consistency~\cite{zhu2017unpaired,kim2017learning}. Liu \textit{et al.}~\cite{liu2017unsupervised} show that unsupervised mappings can be learned by imposing a joint latent space between the encoder and the decoder. In this work we consider the case were paired data is available between some domains or modalities and not available between others (i.e. zero-pair), and how this knowledge can be transfered to those zero-pair cases. In concurrent work, Choi \textit{et al.}~\cite{choi2017stargan} also address scaling to multiple domains (always in the RGB modality) by using a single encoder-decoder model. In contrast, our approach uses multiple cross-aligned encoders and decoders. Our cross-modal setting is also requires deeper structural changes and  modality-specific encoder-decoders.

\paragraph{Multimodal encoder-decoders} Encoder-decoder networks can be extended into multi-way encoder-decoder networks by adding encoders and/or decoders for multiple domains together. Recently, joint encoder-decoder architectures have been used in multi-task settings, where the network is trained to perform multiple tasks (e.g. depth estimation, semantic segmentation, surface normals)~\cite{eigen2015predicting,kendall2017multi}, and multimodal settings, where the inputs data can be from different modalities or even combine several ones~\cite{ngiam2011multimodal}. 

Training a multimodal encoder-decoder network was recently studied in~\cite{Kuga_2017_ICCV}. They use a joint latent representation space for the various modalities. In our work we consider the alignment and transferability of pairwise image translations to unseen translations, rather than joint encoder-decoder architectures. Another multimodal encoder-decoder network was studied in ~\cite{cadena2016multi}. They show that multimodal autoencoders can address the depth estimation and semantic segmentation tasks simultaneously, even in the absence of some of the input modalities. All these works do not consider the zero-pair image translation problem addressed in this paper. 

\paragraph{Zero-shot recognition} In conventional supervised image recognition, the objective is to predict the class label that is provided during training~\cite{lampert2014attribute,fu2017recent}. However, this poses limitations in scalability to new classes, since new training data and annotations are required. In zero-shot learning, the objective is to predict an unknown class for which there is no image available, but a description of the class (i.e. \textit{class prototype}). This description can be a set of attributes(e.g. has wings, blue, four legs, indoor) ~\cite{lampert2014attribute,jayaraman2014zero}, concept ontologies~\cite{fergus2010semantic,rohrbach2011evaluating} or textual descriptions~\cite{reed2016learning}. In general, an intermediate semantic space is leveraged as a bridge between the visual features from seen classes and class description from unseen ones.
In contrast to zero-shot recognition, we focus on unseen translations (unseen input-output pairs rather than simply unseen class labels).

\paragraph{Zero-pair language translation} Evaluating models on unseen language pairs has been studied recently in machine translation~\cite{johnson2016google,chen2017teacher,zheng2017maximum,firat2016multi}. Johnson et al.~\cite{johnson2016google} proposed a neural language model that can translate between multiple languages, even pairs of language where no explicit paired sentences where provided\footnote{Note that~\cite{johnson2016google} refers to this as \textit{zero-shot} translation. In this paper we refer to this setting as zero-pair to emphasize that what is unseen is paired data and avoid ambiguities with traditional zero-shot recognition which typically refers to unseen samples.}. In their method, the encoder, decoder and attention are shared. In our method we focus on images, which are essentially a radically different type of data, with two dimensional structure in contrast to the sequential structure of language.

\section{Encoder-decoder alignment}
\subsection{Multi-domain image translation}
We consider the problem of image translation from domain $\mathcal{X}^{(i)}$ to domain $\mathcal{X}^{(j)}$ as $T_{ij} \colon x^{(i)} \mapsto x^{(j)}$. In our case it is implemented as a encoder-decoder chain $x^{(j)}=T_{ij}\left(x^{(i)}\right)=g_j\left(f_i\left(x^{(i)}\right)\right)$ with encoder $f_i$ and decoder $g_j$ (see Figure~\ref{fig_zero-pair_translation}). The domains connected during training are all trained jointly, and in both directions. It is important to note that for each domain one encoder and one decoder are trained. By training all these encoders and decoders jointly the latent representation is encouraged to align. As a consequence of the alignment of the latent space we can do zero-pair translation at testing time between the domains for which no training pairs were available. The main aim of this article is to analyze to what extend this alignment allows for zero-pair image translation. 

\subsection{Aligning for zero-pair translation}\label{sec:i2imultiway}
Zero-pair translation in images is especially challenging due to the inherent complexity of images, especially in multimodal settings. Ideally, a good latent representation that also works in unseen translations should be not only well-aligned but also unbiased to any particular domain. In addition, the encoder-decoder system should be able to preserve the spatial structure, even in unseen transformations.

\minisection{Autoencoders} One way to improve alignment is by jointly training domain-specific autoencoders with the image translation networks. By sharing the weights between the auto-encoders and the image translation encoder-decoders the latent space is forced to align.

\minisection{Latent space consistency}
The latent space can be enforced to be invariant across multiple domains. Taigman \textit{et al.}~\cite{taigman2016unsupervised} use L2 distance between a latent representation and the reconstructed after another decoding and encoding cycle. When paired samples  $\left(x^{(i)},x^{(j)}\right)$ are available, we propose using cross-domain latent space consistency in order to enforce $f_i\left(x^{(i)}\right)$ and $f_j\left(x^{(j)}\right)$ to be aligned.

\minisection{Preserving spatial structure using side information} In general, image translation tasks result in output images with similar spatial structure as the input ones, such as scene layouts, shapes and contours that are preserved across modalities.
In fact, this spatial structure available in the input image is critical to simplify the problem and achieve good results,
and successful image translation methods often use multi-scale intermediate representations from the encoder as side information to guide the decoder in the upsampling process.
Skip connections are widely used for this purpose. However, conditional decoders using skip connections expect specific information from a particular domain-specific encoder that would be unlikely to work in unseen encoder-decoder pairs. 
Motivated by efficiency, pooling indices~\cite{badrinarayanan2015segnet} were recently proposed as a compact descriptor to guide the decoder in lightweight models. We show here that pooling indices constitute robust and relatively encoder-independent side information, suitable for improving decoding even in unseen translations.

\minisection{Adding noise to latent space} We found that adding some noise at the output of each encoder also helps to train the network and improves the results during test. This seems to help in obtaining more invariance in the common latent representation and better alignment across modalities.

\subsection{Scalable image translation}\label{sec:scale}
One of the advantages of our mix and match networks is that the system can infer many pairwise domain-to-domain translations when the number of domains is high, without explicitly training them. Other pairwise methods where encoders and decoders are not cross-aligned, such as CycleGAN\cite{zhu2017unpaired}, would require training $N\times(N-1)/2$ encoders and decoders for $N$ domains. For mix and match networks each encoder and decoder should be involved in at least one translation pair during training in order to be aligned with the others, thereby reducing complexity from quadratic to linear with the number of domains (i.e. $N-1$ encoders/decoders).

\section{Zero-pair cross-modal image translation}
We propose a challenging cross-modal setting to evaluate zero-pair image translation involving three modalities\footnote{Here the term \textit{modality} has the same role of \textit{domain} in the previous section.}: RGB, depth and semantic segmentation.
It is important to observe that a setting involving heterogeneous modalities\footnote{For simplicity, we will refer to semantic segmentation maps and depth as modalities rather than tasks} (in terms of complexity, number and meaning of different channels, etc.) is likely to require modality-specific architectures and losses.

\subsection{Problem definition}
We consider the problem of jointly learning RGB-to-segmentation translation $y=T_{RS}\left(x\right)$  with and RGB-to-depth translation $z=T_{RD}\left(x\right)$ and evaluating on an unseen transformation $y=T_{DS}\left(z\right)$. The first translation is learned from a semantic segmentation dataset $\mathcal{D}^{(1)}$ with pairs $\left(x,y\right)\in \mathcal{D}^{(1)}$ of RGB images and segmentation maps, and the second from a disjoint RGB-D dataset $\mathcal{D}^{(2)}$ with pairs of RGB and depth images $\left(x,z\right)\in \mathcal{D}^{(2)}$. Therefore no depth image and segmentation map pairs are available to the system. However, note that the RGB images from both datasets could be combined if necessary (we denote this dataset as $\mathcal{X}=\lbrace x \vert x \in \mathcal{D}^{(1)}\cup \mathcal{D}^{(2)} \rbrace $. The system is evaluated on a third dataset $\mathcal{D}^{(3)}$ with paired depth images and segmentation maps. 

\subsection{Mix and match networks architecture}
\begin{figure*}[t]
\centering
\includegraphics[width=\textwidth]{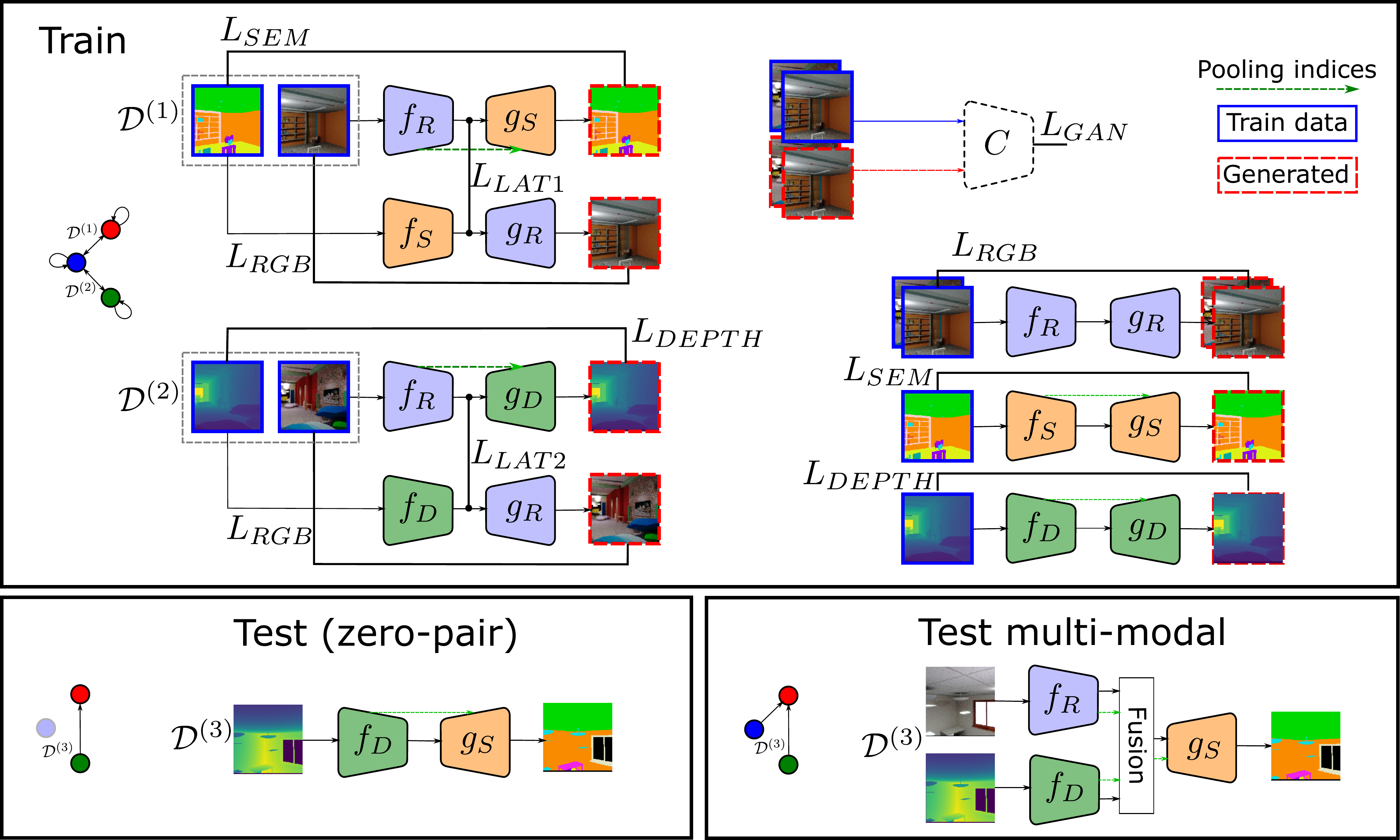}
\caption{Zero-pair cross-modal and multimodal image translation with mix and match networks. Two disjoint sets $\mathcal{D}^{(1)}$ and $\mathcal{D}^{(2)}$ are seen during training, containing (RGB,depth) pairs and (RGB,segmentation) pairs, respectively. The system is tested on the unseen translation depth-to-segmentation (zero-pair) and (RGB+depth)-to-segmentation (multimodal), using a third unseen set $\mathcal{D}^{(3)}$. Encoders and decoders with the same color share weights. Better viewed in color.}
\label{fig_multimodal_framework}
\end{figure*}

The overview of the framework is shown in Fig.~\ref{fig_multimodal_framework}. As basic building blocks we use three modality-specific encoders $f_R\left(x\right)$, $f_D\left(x\right)$ and $f_S\left(x\right)$ (RGB, depth and semantic segmentation, respectively), and the corresponding three modality-specific decoders $g_R\left(h\right)$, $g_D\left(h\right)$ and $g_S\left(h\right)$, where $h$ is the latent representation in the shared space. The required translations are implemented as $y=T_{RS}\left(x\right)=g_S\left(f_R\left(x\right)\right)$, $z=T_{RD}\left(x\right)=g_D\left(f_R\left(x\right)\right)$ and $y=T_{DS}\left(z\right)=g_S\left(f_D\left(z\right)\right)$.

Encoder and decoder weights are shared across the different translations involving same modalities (same color in Fig.~\ref{fig_multimodal_framework}). To enforce better alignment between encoders and decoders of the same modality, we also include self-translations using the corresponding three autoencoders $T_{RR}(x)=g_R\left(f_R\left(x\right)\right)$, $T_{DD}(y)=g_D\left(f_D\left(x\right)\right)$ and $T_{SS}(x)=g_S\left(f_S\left(x\right)\right)$.

We based our encoders and decoders on the SegNet architecture~\cite{badrinarayanan2015segnet}. The encoder of SegNet itself is based on the 13 convolutional layers VGG-16 architecture~\cite{simonyan2014very}. The decoder mirrors the encoder architecture with 13 deconvolutional layers. All encoders and decoders are randomly initialized except for the RGB encoder which is pretrained on ImageNet.

As in SegNet, pooling indices at each downsampling layer of the encoder are provided to the corresponding upsampling layer of the (seen or unseen) decoder\footnote{The RGB decoder does not use pooling indices, since in our experiments we observed undesired grid-like artifacts in the RGB output.}. These pooling indices seem to be relatively similar across the three modalities and effective to transfer spatial structure information that help to obtain better depth and segmentation boundaries in higher resolutions. Thus, they provide relatively modality-independent side information.

\subsection{Loss functions}
As we saw before, a correct cross-alignment between encoders and decoders that have not been connected during training is critical for zero-pair translation. The final loss combines a number of modality-specific losses for both cross-domain translation and self-translation (i.e. autoencoders) and alignment constraints in the latent space
\begin{equation*}
L = \lambda_R L_{RGB}+\lambda_S L_{SEG}+\lambda_D L_{DEPTH}+\lambda_A L_{LAT}
\end{equation*}

\paragraph{RGB-specific} We use a combination of L2 distance and adversarial loss $L_{RGB}=\lambda_{L2} L_{L2}+ L_{GAN}$. L2 distance is used to compare the estimated and the ground truth RGB image after translation from a corresponding depth or segmentation image. It is also used in the RGB autoencoder
\begin{eqnarray*}
L_{L2} &=&  \mathbb{E}_{(x,y)\sim p_{\mathcal{D}^{(1)}}(x,y)}\left[ \left\| T_{SR}\left(y\right) - x  \right\|_2 \right] \\
       &+& \mathbb{E}_{(x,z)\sim p_{\mathcal{D}^{(2)}}(x,z)}\left[ \left\| T_{DR}\left(z\right) -x \right\|_2 \right] \\
      &+& \mathbb{E}_{x\sim p_{\mathcal{X}(x)}}\left[ \left\| T_{RR}\left(x\right) -x \right\|_2 \right]
\end{eqnarray*}
In addition, we also include the least squares adversarial loss \cite{mao2016multi,isola2016image} on the output of the RGB decoder
\begin{equation*}
L_{GAN} = \mathbb{E}_{x\sim p_{\mathcal{X}}(x)}\left[ \left( C\left(x\right) -1 \right)^2 \right]+ \mathbb{E}_{\hat{x}\sim \hat{p}(x)}\left[ \left( C\left(\hat{x}\right) \right)^2 \right]
\end{equation*}
where $\hat{p}(x)$ is the resulting distribution of the combined images $\hat{x}$ generated by $\hat{x}=T_{SR}\left(y\right)$, $\hat{x}=T_{DR}\left(z\right)$ and $\hat{x}=T_{RR}\left(x\right)$. Note that the RGB autoencoder and the discriminator $C\left(x\right)$ are both trained with the combined RGB data $\mathcal{X}$.

\paragraph{Depth}
For depth we use the Berhu loss ~\cite{laina2016deeper} in both RGB-to-depth translation and in the depth autoencoder
\begin{eqnarray*}
L_{DEPTH} &=& \mathbb{E}_{(x,z)\sim p_{\mathcal{D}^{(2)}}(x,z)}\left[ \mathcal{B}\left( T_{RD}\left(x\right) -z \right) \right] \\
      	  &+& \mathbb{E}_{(x,z)\sim p_{\mathcal{D}^{(2)}}(x,z)}\left[ \mathcal{B}\left( T_{DD}\left(z\right) -z \right) \right]
\end{eqnarray*}
where $\mathcal{B}\left(z\right)$ is the average Berhu loss.

\paragraph{Semantic segmentation} For segmentation we use the average cross-entropy loss $\mathcal{CE}\left(\hat{y},y\right)$ in both RGB-to-segmentation translation and the segmentation autoencoder
\begin{eqnarray*}
L_{SEM} &=& \mathbb{E}_{(x,y)\sim p_{\mathcal{D}^{(1)}}(x,y)}\left[ \mathcal{CE}\left( T_{RS}\left(x\right),y \right) \right] \\
        &+& \mathbb{E}_{(x,y)\sim p_{\mathcal{D}^{(2)}}(x,y)}\left[ \mathcal{CE}\left( T_{SS}\left(y\right),y \right) \right].
\end{eqnarray*}

\paragraph{Latent space consistency} We enforce latent representations to remain close independently of the encoder that generated them. In our case we have two consistency losses

\begin{eqnarray*}
L_{LAT} &=& L_{LAT1}+ L_{LAT2} \\
L_{LAT1} &=& \mathbb{E}_{(x,y)\sim p_{\mathcal{D}^{(1)}}(x,y)}\left[ \left\| f_R\left(x\right) - f_S\left(y\right) \right\|_2 \right] \\
		L_{LAT2} &=& \mathbb{E}_{(x,z)\sim p_{\mathcal{D}^{(2)}}(x,z)}\left[ \left\| f_R\left(x\right) - f_D\left(z\right) \right\|_2 \right]        
\end{eqnarray*}

\section{Experimental Results}
To the best of our knowledge there is no existing work which reports results for the setting of zero-pair image translation. In particular, we evaluate the proposed mix and match networks on zero-pair translation for semantic segmentation from depth images (and viceversa), and we show results for semantic segmentation from multimodal data. Finally, we illustrate the possibility to perform zero-pair translations for unpaired datasets, and the advantage of mix and match networks in terms of scalability. 

\subsection{Datasets and experimental settings}

\minisection{SceneNetRGBD}  The SceneNetRGBD dataset~\cite{McCormac:etal:ICCV2017} consists of 16865 synthesized train videos and 1000 test videos. Each of them includes 300 matching triplets (RGB, depth and segmentation map), with a size of 320x240 pixels. In our examples, we use two subsets as our datasets: 

\begin{itemize}
	\item 51K dataset: the train set is selected from the first 50 frames from each of the first 1000 videos in the train set. The test set is collected by selecting the 60th frame from the same 1000 videos. This dataset was used to evaluate some of the architecture design choices.
  
    \item 170K dataset: We collected a larger dataset which consists of 150K triplets for the train set, 10K triplets for the validation set and 10K triplets for the test set. The 10K validation set is also from the train set of SceneNetRGBD. For the train set, we select 10 triplets from the first 150000 training triplets. The triplets are sampled from the first frame to last frame every 30 frames. The validation set is from the remaining videos of the train set and the test set is taken from the test dataset.
\end{itemize}

Each train set is divided into two equal non-overlapping splits from different videos. Although the collected splits contain triplets, we only use pairs to train our model.
 
Following common practices in these tasks, for segmentation we compute the intersection over union (IoU) and report per-class average (mIoU), and the global scores, which gives the percentage of correctly classified pixels. For depth we also include quantitative evaluation, following the standard error metrics for depth estimation~\cite{eigen2015predicting}.

\minisection{Network training} We use  Adam~\cite{kingma2014adam} with batch size of 6, using a learning rate of 0.0002. We set $\lambda_R = 1$, $\lambda_S = 100$, $\lambda_D = 10$, $\lambda_A =1$, $\lambda_{L2} =1$. For the first 200,000 iterations we train all networks. For the following 200,000 iterations we use $\lambda_R = 10$, $\lambda_A =10$, $\lambda_{L2} =10$ and  freeze the RGB encoder. We found the network converges faster using a large initial $\lambda_A$.  
We add Gaussian noise to the latent space with zero mean and a standard deviation of 0.5.
 
\begin{table}[tb]
\centering
\setlength{\arrayrulewidth}{1.8\arrayrulewidth}
\begin{tabular}{cc|cc}
\hline
Side information  &Pretrained  &mIoU  &Global\;
\\ 
\hline  
-  & N  &{32.2\%}    &{63.5\%}      \\ 
Skip connections  & N     &{14.1\%}    &{52.6\%}      \\
Pooling indices  & N  &{45.6\%}    &{73.4\%}      \\
Pooling indices  & Y    &{49.5\%}    &{80.0\%}      \\
\hline
\end{tabular}
\caption{Influence of side information and RGB encoder pretraining on the final results. The task is zero-shot depth-to-semantic segmentation.}
\label{table:connections}
\end{table}

\begin{table}[tb]
\centering
\setlength{\arrayrulewidth}{1.8\arrayrulewidth}
\begin{tabular}{ccc|cc}
\hline
AutoEnc   &Latent   &Noise &mIoU  &Global
\\ 
\hline  
 N  & N  & N   &{5.64\%}    &{13.5\%}   \\
 Y  & N  & N   &{22.9\%}    &{52.6\%}   \\
 Y  & Y  & N   &{48.9\%}    &{78.2\%}  \\
 Y  & Y  & Y   &{49.5\%}    &{80.0\%}  \\
\hline
\end{tabular}
\caption{Impact of several components (autoencoder, latent space consistency loss and noise) in the performance. The task is zero-shot depth-to-semantic segmentation.}
\label{table:ablation}
\end{table}

\subsection{Ablation study}
In a first experiment we use the 51K dataset to study the impact of several design elements on the overall performance of the system. 

\begin{figure}[t]
\centering
\includegraphics[width=0.8\columnwidth]{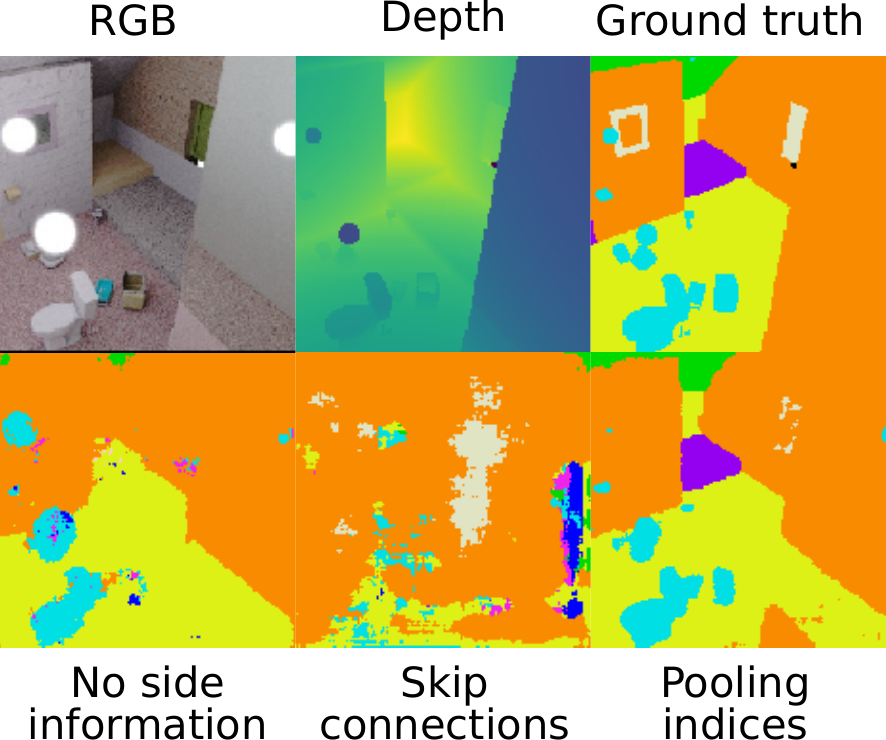}
\caption{\label{fig:example_depth-to-segm} Role of  side information in unseen depth-to-segmentation translation.}
\end{figure}

\minisection{Side information} 
We first evaluate the usage of side information from the encoder to guide the upsampling process in the decoder. We consider three variants: no side information, skip connections~\cite{ronneberger2015u} and pooling indices~\cite{badrinarayanan2015segnet}. The results in Table~\ref{table:connections} show that skip connections obtain worse results than no side information at all. This is due to the fact that skip connections are not domain-invariant and at testing time when we combine an encoder and decoder these connections result in a different input from the one seen under training, resulting in a drop of performance.
Fig.~\ref{fig:example_depth-to-segm} illustrates the differences between these three variants. Without side information the network is able to reconstruct a coarse segmentation but without further guidance it is not able to refine it properly. Skip connections provide features that could guide the decoder but instead confuse it, since in the zero-pair case the decoder has not seen the features of that particular encoder. Pooling indices are more invariant as side information and obtaining the best results.

\minisection{RGB pretraining} We also compare training the RGB encoder from scratch and initializing with pretrained weights from ImageNet. Table~\ref{table:connections} show an additional gain of around 5\% in mIoU when using the pretrained weights.

Given these results we perform all the remaining experiments initializing the RGB encoder with pretrained weights and use pooling indices as side information.

\minisection{Latent space consistency, noise and autoencoders} We evaluate these three factors, with Table~\ref{table:ablation} showing that latent space consistency and the usage of autoencoders lead to significant performance gains; for both, the performance (in mIoU) is more than doubled. Adding noise to the output of the encoder results in a small performance gain. 

\begin{table*}[tb]
\setlength{\tabcolsep}{2.5pt}
\centering
\resizebox{\textwidth}{!}{
\begin{tabular}{ccc|ccccccccccccc|cc}
\hline
Method & Conn. & $L_{SEM}$   & \rotatebox{90}{Bed}  &\rotatebox{90}{Book\;}   &\rotatebox{90}{Ceiling\;}   &\rotatebox{90}{Chair\;}   &\rotatebox{90}{Floor}  &\rotatebox{90}{Furniture}  &\rotatebox{90}{Object\;}   &\rotatebox{90}{Picture\;}  &\rotatebox{90}{Sofa\;} &\rotatebox{90}{Table\;} &\rotatebox{90}{TV\;} &\rotatebox{90}{Wall\;}   &\rotatebox{90}{Window\;}   
&\rotatebox{90}{mIoU\;}  &\rotatebox{90}{Global\;} 
\\
\hline
\textbf{Baselines} & & & & & & & & & & &&&&&&& \\
CycleGAN~\cite{zhu2017unpaired}
& SC & CE &{2.79}    &{0.00}   &{16.9}  &{6.81}    &{4.48}    &{0.92}    &{7.43}   &{0.57}  &{9.48}    &{0.92}     &{0.31}    &{17.4}   &{15.1}  &{6.34}     &{14.2}   \\ 
2$\times$pix2pix~\cite{isola2016image}
&SC &CE &34.6 &1.88 &70.9 &20.9 &63.6 &17.6 &14.1 &0.03 &38.4 &10.0 &4.33 &67.7 &20.5  &25.4 &57.6
\\
M\&MNet $D\rightarrow R \rightarrow S$
& PI & CE &0.02 &0.00 &8.76 &0.10 &2.91 &2.06 &1.65 &0.19 &0.02 &0.28 &0.02 &58.2 &3.3 &5.96 &32.3 
\\ 
M\&MNet $D\rightarrow R \rightarrow S$
& SC & CE &25.4 &0.26 &82.7 &0.44 &56.6 &6.30 &23.6 &5.42 &0.54 &21.9 &10.0 &68.6 &19.6 &24.7 &59.7 
\\ 
\hline
\textbf{Zero-pair} & & & & & & & & &&&&&&& \\
M\&MNet $D \rightarrow S$
& PI & CE &{50.8}    &{18.9}   &{89.8}  &{31.6}   &{88.7}    &{48.3}    &{44.9}   &{62.1}  &{17.8}    &{49.9}     &{51.9}    &{86.2}   &{79.2} &55.4    &80.4
\\ 
\hline
\textbf{Multi-modal} & & & & & & & & &&&&&&& \\
M\&MNet $\left( R,D\right) \rightarrow S$
& PI & CE &{49.9}    &{25.5}   &{88.2}  &{31.8}    &{86.8}    &{56.0}    &{45.4}   &{70.5}  &{17.4}    &{46.2}     &{57.3}    &{87.9}   &{79.8}  &{57.1}     &{81.2}   
\\ 
\hline
\end{tabular}
}
\caption{Zero-pair depth-to-semantic segmentation. \textbf{SC}: skip connections, \textbf{PI}: pooling indexes, \textbf{CE}: cross-entropy}
\label{table:depth2segm}
\end{table*}

\begin{table}
\setlength{\tabcolsep}{2.5pt}
\resizebox{\columnwidth}{!}{
\begin{tabular}{c|ccccc}
\hline
\multirow{2}{*}{Method} & \multicolumn{3}{c}{$\delta<$} &RMSE &RMSE \\
 & $1.25$ & $1.25^2$ & $1.25^3$ &(lin) &(log) \\
\hline
\textbf{Baselines} & & & & & \\
CycleGAN~\cite{zhu2017unpaired} &{0.05}    &{0.12}   &{0.20}  &{4.63}    &{1.98}       \\ 
M\&MNet $S\rightarrow R \rightarrow D$ &{0.15}    &{0.30}   &{0.44}  &{3.24}    &{1.24}       \\
\hline
\textbf{Zero-pair} & & & & & \\
M\&MNet $S \rightarrow D$ &{0.33}    &{0.42}   &{0.59}  &{2.8}    &{0.67}       \\
\hline
\textbf{Multi-modal} & & & & & \\
M\&MNet $\left( R,S\right) \rightarrow D$
&{0.36}    &{0.48}   &{0.65}  &{2.48}    &{0.64}       \\
\hline
\end{tabular}
}
  \caption{Zero-pair semantic segmentation-to-depth.}\label{table:segm2depth}%
\end{table}

\subsection{Comparison with other methods}
We compare the results of our mix and match networks for depth to segmentation, $D \rightarrow S$, to the following two baselines:

\begin{itemize}
\item CycleGAN~\cite{zhu2017unpaired} learns a mapping from depth to semantic segmentation without explicit pairs. In contrast to ours, this method only leverages depth and semantic segmentation, ignoring the available RGB data and the corresponding pairs.
\item 2$\times$pix2pix~\cite{isola2016image} learns from paired data two encoder-decoder pairs ($D \rightarrow R$ and $R \rightarrow S$). The architecture uses skip connections and the corresponding modality-specific losses. We use the exact code from~\cite{isola2016image}. In contrast to ours, it requires explicit decoding to RGB, which may degrade the quality of the prediction.
\item $D \rightarrow R \rightarrow S$ is similar as the 2$\times$pix2pix but than with a similar architecture as we use in our M\&MNet. We train a translation model from depth to RGB and from RGB to segmentation, and obtain the transformation depth-to-segmentation by concatenating them. Note that it requires using an intermediate RGB image.
\end{itemize}

Table~\ref{table:depth2segm} compares the three methods on the 170K dataset. CycleGAN is not able to learn a good mapping from depth to semantic segmentation, showing the difficulty of unpaired translation to solve this complex cross-modal task. 2$\times$pix2pix manages to improve the results by resorting to the anchor domain RGB, although still not satisfactory since the first translation network drops information not relevant for the RGB task but necessary for reconstructing depth (like in the ”Chinese whispers”/telephone game). 

Mix and match networks evaluated on ($D \rightarrow R \rightarrow S$) achieve a similar result to CycleGAN, but significantly worse than 2$\times$pix2pix. However, when we run our architecture with skip connections we obtain similar results as 2$\times$pix2pix. Note that because in this setting the encoders and decoders are used in the same setting in both training and testing, skip connections function well.  

\begin{figure}[t]
    \centering
    \begin{subfigure}[b]{\columnwidth}
          \centering
          \begin{subfigure}[b]{0.3\columnwidth}
                  \centering
                  \includegraphics[width=\columnwidth]{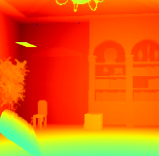}
                  \caption{Input: depth}
          \end{subfigure}%
          \begin{subfigure}[b]{0.3\columnwidth}
                  \centering
                  \includegraphics[width=\columnwidth]{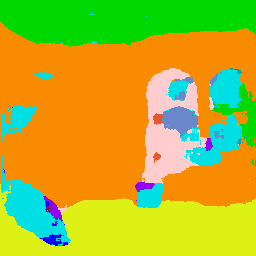}
                  \caption{$2\times$Pix2pix}
          \end{subfigure}
          \begin{subfigure}[b]{0.3\columnwidth}
                  \centering
                  \includegraphics[width=\columnwidth]{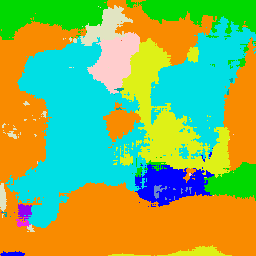}
                  \caption{CycleGAN}
          \end{subfigure}
    \end{subfigure}
    \begin{subfigure}[b]{\columnwidth}
          \centering
          \begin{subfigure}[b]{0.3\columnwidth}
                  \centering
                  \includegraphics[width=\columnwidth]{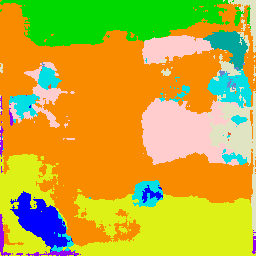}
                  \caption{D$\rightarrow$R$\rightarrow$S}
          \end{subfigure}%
          \begin{subfigure}[b]{0.3\columnwidth}
                  \centering
                  \includegraphics[width=\columnwidth]{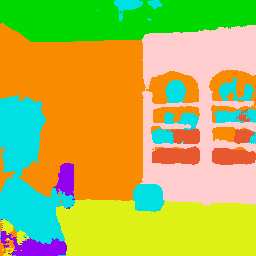}
                  \caption{Proposed}
          \end{subfigure}
          \begin{subfigure}[b]{0.3\columnwidth}
                  \centering
                  \includegraphics[width=\columnwidth]{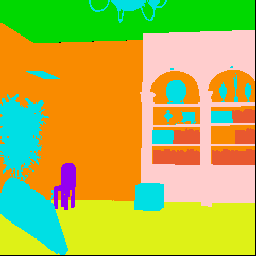}
                  \caption{Ground truth}
          \end{subfigure}
    \end{subfigure}
        \caption{\label{fig:rgbds_comparison} Different methods evaluated on zero-pair depth$\rightarrow$segmentation.}
\end{figure}

The direct combination  ($D \rightarrow S$) outperforms all baselines significantly. The results more than double in terms of mIoU. Figure~\ref{fig:rgbds_comparison} illustrates the comparison between our approach and the baselines; our method is the only one that manages to identify the main semantic classes and their general contours in the image. In conclusion, the results show that mix and match networks enable effective zero-pair translation. 

In Table~\ref{table:segm2depth} we show the results when we test in the opposite direction from semantic segmentation to depth. The conclusions are similar as in previous experiment. Again our method $S \rightarrow D$ outperforms both baseline methods on all five evaluation metrics. Fig.~\ref{fig:example_segm-to-depth} illustrates this case, showing how pooling indices are also key to obtain good depth images, compared with no side information at all.

\subsection{Multimodal translation}
\begin{figure}[tb]
\centering
\includegraphics[width=0.8\columnwidth]{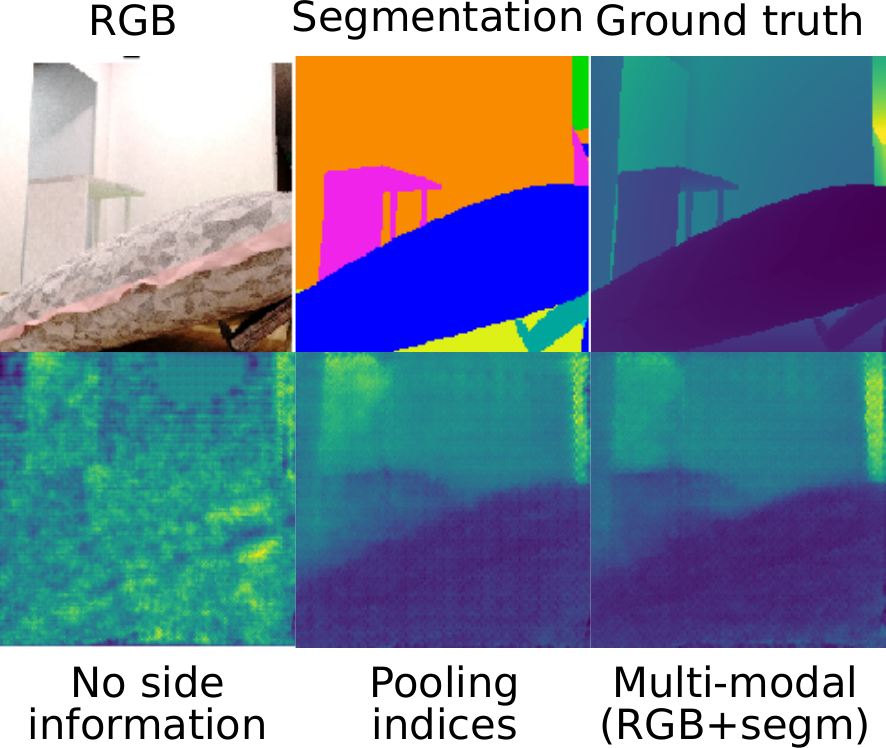}
\caption{\label{fig:example_segm-to-depth} 
Zero-pair and multimodal segmentation-to-depth.}
\end{figure}

Next we consider the case of multimodal translation from pairs (RGB, depth) to semantic segmentation. As depicted in Figure~\ref{fig_multimodal_framework} multiple modalities can be combined (since the latent spaces are aligned) at the input of  semantic segmentation decoder. To combine the two modalities we perform a weighted average of both RGB and depth latent vectors (the weight $\alpha$ ranges from $\alpha=0$, only RGB, and $\alpha=1$, only depth, depending on the case). We set $\alpha$ to 0.2 and use the pooling indices from the RGB encoder (instead of those from the  depth encoder, see supplementary material for more details). The results in Table~\ref{table:depth2segm} and Table~\ref{table:segm2depth} and the example in Figure~\ref{fig:example_segm-to-depth} show that this multimodal combination further improves the performance of zero-pair translation. 
 
\subsection{Scalable unpaired image translation}
  
\begin{figure}[tb]
\centering
\begin{subfigure}[t]{\columnwidth}
		\includegraphics[width=0.99\columnwidth]{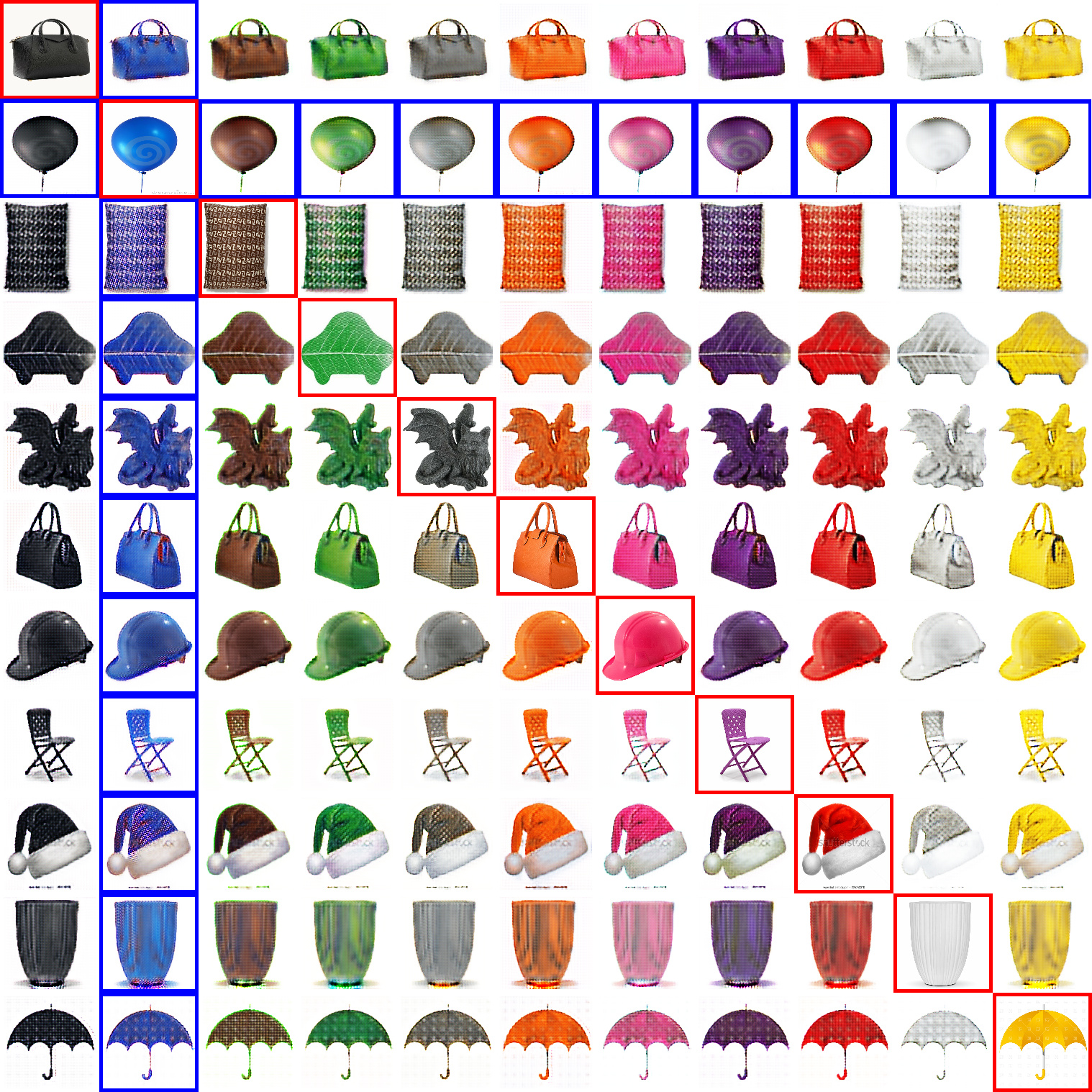}
\caption{Color transfer: only transformations from or to blue (anchor domain) are seen. Input images are highlighted in red and seen translations in blue.\label{fig:color_results}}
\end{subfigure}

\begin{subfigure}[t]{\columnwidth}
\includegraphics[width=\columnwidth]{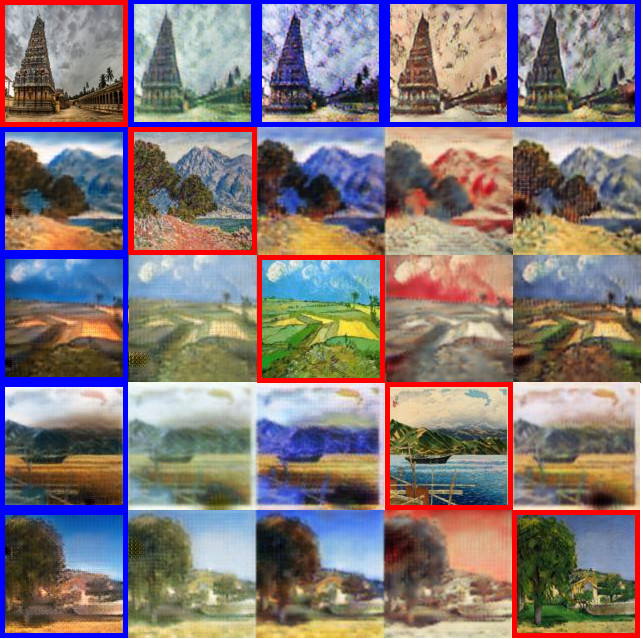}
\caption{\label{fig:style_transfer}Style transfer: trained on four styles + photo (anchor) with data from \cite{zhu2017unpaired}). From left to right: photo, Monet, van Gogh, Ukiyo-e and Cezanne. Input images are highlighted in red and seen translations in blue.}
\end{subfigure}
\caption{\label{fig:cross-domain}Zero-pair cross-domain unpaired translations.}
\end{figure}

As explained in Section~\ref{sec:scale}, mix and match networks scale linearly with the number of domains, whereas existing unpaired image translation methods scale quadratically. As examples of translations between many domains, we show results for object recoloring and style transfer, using mix and match networks based on multiple CycleGANs~\cite{zhu2017unpaired} combined with autoencoders. For the former we use the colored objects dataset~\cite{yu2018beyond} with eleven distinct colors ($N=11$) and around 1000 images per color. Covering all possible image-to-image recoloring combinations requires training a total of $N\left(N-1\right)/2=55$ encoders (and decoders) using CycleGANs. In contrast, mix and match networks only require to train $N-1=10$ encoders and eleven decoders, while still successfully addressing the recoloring task (see Figure~\ref{fig:color_results}).
Similarly, scalable style transfer can be addressed using mix and match networks (see Figure~\ref{fig:style_transfer} for an example).

\section{Conclusion}
In this paper we introduce the problem of zero-pair image translation, where knowledge learned in paired translation models can be effectively transferred and leveraged to perform new unseen translations. The image-to-image scenario poses several challenges to the alignment of encoders and decoders in a way that guarantees cross-domain transferability and without too much dependence on the domain or the modality. We studied this scenario in zero-pair cross-modal and multimodal settings. Notably, we found that side information in the form of pooling indices is robust to modality changes and very helpful to guide the reconstruction of spatial structure. Other helpful techniques are cross-modal consistency losses and adding noise to the latent representation.

\minisection{Acknowledgements}
Herranz acknowledges the European Union’s H2020 research under Marie Sklodowska-Curie grant No. 6655919. We acknowledge the   project TIN2016-79717-R, the CHISTERA project M2CR (PCIN-2015-251) of the Spanish Government and the CERCA Programme of Generalitat de Catalunya. Yaxing Wang acknowledges the Chinese Scholarship Council (CSC) grant No.201507040048. We also acknowledge the generous GPU support from Nvidia.

\bibliographystyle{ieee}
\bibliography{refs}


\appendix

\section{Appendix: Network architecture}
Table~\ref{table:encoders} shows the architecture (convolutional and pooling layers) of the encoders used in the cross-modal experiment. Tables~\ref{table:decoders} and \ref{table:rgb_decoder} show the corresponding decoders. Table~\ref{table:rgb_discriminator} shows the discriminator used for RGB. . Every convolutional layer of the encoders, decoders and the discriminator is followed by a batch normalization layer and a ReLU layer (LeakyReLU for the discriminator). The only exception is the RGB encoder, which is is initialized with weights from the VGG16 model\cite{simonyan2014very} and does not use batch normalization.

\begin{table}[h]
\centering
\setlength{\arrayrulewidth}{0.4\arrayrulewidth}
\resizebox{\columnwidth}{!}{
\begin{tabular}{cc|cc}
\hline
layer  &Input $\rightarrow $Output    &Kernel, stride \\
\hline
conv1    & [6,8,8,512] $\rightarrow [6,16,16,512]$ & [3, 3], 1\\ 
conv2    & [6,16,16,512] $\rightarrow [6,32,32,256]$ & [3, 3], 1\\ 
conv3    & [6,32,32,256] $\rightarrow [6,64,64,128]$ & [3, 3], 1\\ 
conv4    & [6,64,64,128] $\rightarrow [6,128,128,64]$ & [3, 3], 1\\ 
conv5    & [6,128,128,64]$ \rightarrow [6,256,256,3]$ & [3, 3], 1\\ 
\hline
\hline
\end{tabular}
}
\caption{Convolutional and pooling layers of the RGB decoder.}
\label{table:rgb_decoder}
\end{table}

\begin{table}[h]
\centering
\setlength{\arrayrulewidth}{0.4\arrayrulewidth}
\resizebox{\columnwidth}{!}{
\begin{tabular}{cc|cc}
\hline
layer  &Input $\rightarrow $Output    &Kernel, stride \\
\hline
deconv1    &$ [6,256,256,3] \rightarrow [6,128,128,64]$ & [5, 5], 2\\ 
deconv2     &$ [6,128,128,64] \rightarrow [6,64,64,128]$ & [5, 5], 2\\ 
deconv3     &$ [6,64,64,128] \rightarrow [6,32,32,256]$ & [5,5], 2\\ 
deconv4     &$ [6,32,32,256] \rightarrow [6,16,16,512]$ & [5,5], 2\\ 
\hline
\hline
\end{tabular}
}
\caption{RGB discriminator.}
\label{table:rgb_discriminator}
\end{table}

\section{Appendix: Multimodal fusion}

\begin{figure}[h]
\includegraphics[width=\columnwidth]{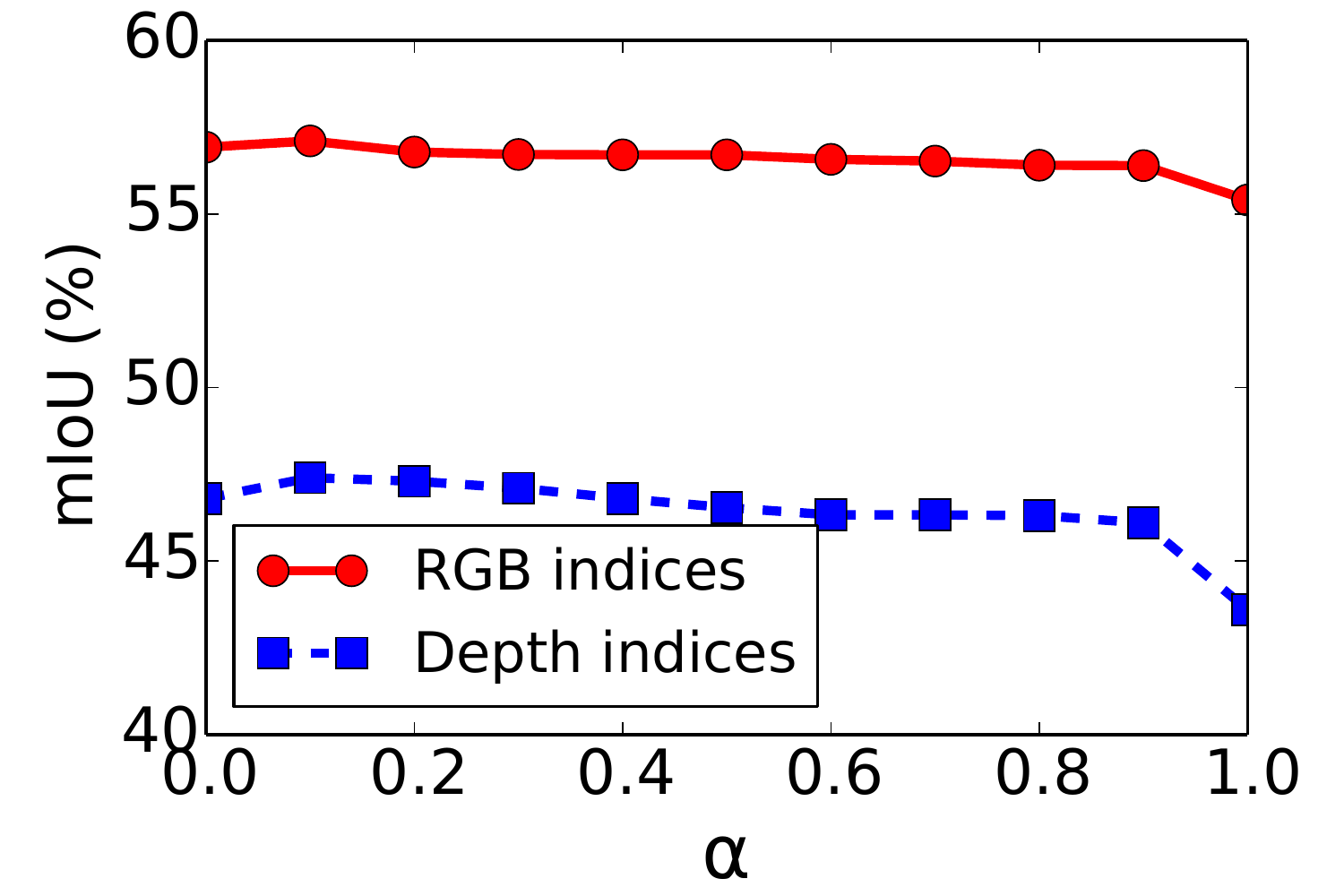}
\caption{Multimodal semantic segmentation: pooling indices modality and modality weight $\alpha$ ($\alpha=0$ for RGB only, $\alpha=1$ for depth only).\label{fig:multimodal}}
\end{figure}

Figure~\ref{fig:multimodal} shows the performance for different values of $\alpha$ for multimodal semantic segmentation. It also compares the performance when the semantic segmentation decoder uses the pooling indices from the depth encoder instead of the ones from the RGB encoder.

\begin{table*}
\centering
\setlength{\arrayrulewidth}{1\arrayrulewidth}
\begin{tabular}{cc|cc}
\hline
Layer &Input $\rightarrow $Output    &Kernel, stride\\ 
\hline  
conv1 (RGB)   & [6,256,256,3] $\rightarrow$ [6,256,256,64] & [3,3], 1\\ 
conv1 (Depth)  & $[6,256,256,1] \rightarrow [6,256,256,64]$ & [3,3], 1 \\
conv1 (Segm.)  &  [6,256,256,14] $\rightarrow$ [6,256,256,64]& [3,3], 1 \\
conv2     & [6,256,256,64] $\rightarrow$ [6,256,256,64] & [3,3], 1\\ 
pool2 (max) & [6,256,256,64] $\rightarrow$ [6,128,128,64]+indices2& [2,2], 2 \\
\hline
conv3     & [6,128,128,64] $\rightarrow$ [6,128,128,128] & [3,3], 1\\ 
conv4     & [6,128,128,128] $\rightarrow$ [6,128,128,128] & [3,3], 1\\ 
pool4 (max) & [6,128,128,128] $\rightarrow$ [6,64,64,128]+indices4 & [2,2], 2 \\
\hline
conv5    & [6,64,64,128] $\rightarrow$ [6,64,64,256] & [3,3], 1\\ 
conv6     & [6,64,64,256] $\rightarrow$ [6,64,64,256] & [3,3], 1\\ 
conv7     & [6,64,64,256] $\rightarrow$ [6,64,64,256] & [3,3], 1\\ 
pool7 (max) & [6,64,64,256] $\rightarrow$ [6,32,32,256]+indices7& [2,2], 2 \\
\hline
conv8     &  [6,32,32,256] $\rightarrow$ [6,32,32,512] & [3,3], 1\\ 
conv9     & [6,32,32,512] $\rightarrow$ [6,32,32,512] & [3,3], 1\\ 
con10     & [6,32,32,512] $\rightarrow$ [6,32,32,512] & [3,3], 1\\ 
pool10 (max) & [6,32,32,512] $\rightarrow$ [6,16,16,512]+indices10& [2,2], 2 \\
\hline
conv11     & [6,16,16,512] $\rightarrow$ [6,16,16,512] & [3,3], 1\\ 
conv12     & [6,16,16,512] $\rightarrow$ [6,16,16,512] & [3,3], 1\\ 
conv13     & [6,16,16,512] $\rightarrow$ [6,16,16,512] & [3,3], 1\\ 
pool13 (max) & [6,16,16,512] $\rightarrow$ [6,8,8,512]+indices13& [2,2], 2 \\
\hline
\end{tabular}
\caption{Convolutional and pooling layers of the encoders.}
\label{table:encoders}
\end{table*}

\begin{table*}
\centering
\setlength{\arrayrulewidth}{1\arrayrulewidth}
\begin{tabular}{cc|cc}
\hline
layer  &Input $\rightarrow $Output    &Kernel, stride\\ 
\hline
unpool1      &indices13 + [6,8,8,512] $\rightarrow [6,16,16,512] $ & [2, 2], 2 \\
conv1     & [6,16,16,512] $\rightarrow [6,16,16,512]$ & [3,3], 1\\ 
conv2     & [6,16,16,512] $\rightarrow [6,16,16,512]$ & [3,3], 1\\ 
conv3     & [6,16,16,512] $\rightarrow [6,16,16,512]$ & [3,3], 1\\ 
\hline
unpool4  &indices10 + [6,16,16,512] $\rightarrow [6,32,32,512]$& [2, 2], 2 \\
conv4     & [6,32,32,512] $\rightarrow [6,32,32,512]$ & [3,3], 1\\ 
conv5    & [6,32,32,512] $\rightarrow [6,32,32,512]$ & [3,3], 1\\ 
conv6     & [6,32,32,512] $\rightarrow [6,32,32,256]$ & [3,3], 1\\ 
\hline
unpool7  & indices7 + [6,32,32,256] $\rightarrow [6,64,64,256]$& [2, 2], 2 \\
conv7     & [6,64,64,256] $\rightarrow [6,64,64,256]$ & [3,3], 1\\ 
conv8     &  [6,64,64,256]$\rightarrow [6,64,64,256]$ & [3,3], 1\\ 
conv9    & [6,64,64,256]$\rightarrow [6,64,64,128]$ & [3,3], 1\\ 
\hline
unpool10  &indices4 + [6,64,64,128] $\rightarrow [6,128,128,128]$& [2, 2], 2 \\
conv10     & [6,128,128,128] $\rightarrow [6,128,128,128]$ & [3,3], 1\\ 
conv11     & [6,128,128,128] $\rightarrow [6,128,128,64]$ & [3,3], 1\\ 
\hline
unpool12  & indices2 + [6,128,128,64] $\rightarrow [6,256,256,64]$& [2, 2], 2 \\
conv12     & [6,256,256,64] $\rightarrow [6,256,256,64]$ & [3,3], 1\\ 
conv13 (Depth)     & [6,256,256,64] $\rightarrow [6,256,256,1]$ & [3,3], 1\\ 
conv13 (Segm.)  & [6,256,256,64] $\rightarrow [6,256,256,14]$ & [3,3], 1\\ 

\hline
\end{tabular}
\caption{Convolutional and pooling layers of the segmentation and depth decoders.}
\label{table:decoders}
\end{table*}

\end{document}